\documentclass{amcs-preprint}

\usepackage{booktabs}   % better lines in tables
\usepackage{multirow}   % multi-row cells in tables
\usepackage{multicol}   % multi-column cells in tables
\usepackage{bbm}        % indicator function
\usepackage{dsfont}     % indicator function

\title[Alike Parts: A Feature-Informed Approach to Local and Global Prototype Explanations]{Alike Parts: A Feature-Informed Approach to Local and Global Prototype Explanations}

\author[ad1][]{Jacek Karolczak}
\author[ad1][]{Jerzy Stefanowski}

\correspondingauthor{Jacek Karolczak}

\address[ad1]{Institute of Computing Science\\ Poznan University of Technology\\ ul. Piotrowo 2, 60-965 Poznań, Poland\\ e-mail: \{\href{mailto:jacek.karolczak@cs.put.poznan.pl}{jacek.karolczak}, \href{mailto:jerzy.stefanowski@cs.put.poznan.pl}{jerzy.stefanowski}\}@cs.put.poznan.pl}

\Runauthors{J. Karolczak and J. Stefanowski}

\usepackage{fancyhdr}
\usepackage{xcolor} % Required for defining and using colors

\definecolor{headergray}{HTML}{888888}
\begin{document}

\begin{abstract}
Prototype-based explanations offer an intuitive, example-based approach to support the interpretability of machine learning black box classifiers but often lack feature-level granularity. We introduce a framework that integrates feature importance at two levels to address this gap. First, for local explanations, we propose \textit{alike parts}: a method that uses feature importance scores to highlight the most relevant, shared feature subsets between a classified instance and its nearest prototype, guiding user attention. Second, we augment the global prototype selection objective function with a feature importance term to actively promote diversity in the feature attributions of the selected prototypes. Experiments on six benchmark datasets show that this augmented selection process maintains or, in some cases, increases the prediction fidelity of the surrogate model, suggesting that feature diversity does not compromise model fidelity.
\end{abstract}

\maketitle
\thispagestyle{fancy}
\section{Introduction}

The growing application of modern artificial intelligence (AI), particularly machine learning (ML), to high-stakes real-world problems, especially those that impact human welfare or critical systems, necessitates requirements beyond predictive accuracy. A primary concern is establishing human trust in the safe and appropriate use of these systems. This need has led to the formalization of the principles known as Trustworthy AI \cite{TrustAI}. Within this framework, explainability has been identified as a cornerstone requirement \cite{TrustXAI}. Note that many of the highest-performing ML methods are black-box models. These models inherently lack transparency, failing to provide insight into the rationale behind their decisions or the internal logic driving their decisions. This fundamental conflict between the critical need for transparency and the opaque nature of leading models is the main driver for the growing field of Explainable Artificial Intelligence (xAI).

The field of XAI is a broad and active research area focused on developing methods to explain black-box models. These methods offer various types of explanation and are mainly categorized by their scope. \textit{Local explanations} provide the reasons for a system's decision on a single specific instance. In contrast, \textit{global explanations} operate on a larger scale, either by covering a large set of examples or by attempting to approximate the overall internal logic of the ML model \cite{bodria2023benchmarking}.

Among various explanation techniques, such as feature importance~\cite{ribeiro2016lime,lundberg2017shap} and counterfactuals \cite{stepka2024multi}, prototypes are particularly noteworthy. Since prototypes are actual examples of training data, they are inherently easier for humans to understand compared to more abstract explanation methods \cite{mastromichalakis2024prototypes}. Prototypes can serve a dual role: they can function as a \textit{local explanation} by linking a model's prediction to the most similar training instances, or as a \textit{global explanation} by using a limited number of representative examples to approximate the model's overall decision boundaries.

Although prototypes can be applied to various data types, this paper focuses on tabular data, where examples are represented as vectors of $\left(\text{feature}\,,\,\text{value}\right)$ pairs. However, interpreting prototypes remains challenging with many features \cite{mastromichalakis2024prototypes}. For local explanations, users may struggle to determine which features are the most important for a specific prediction. For global explanations, it is desirable that the discovered prototypes are not only well-distributed across the data space but also represent diverse subsets of important feature values. This goal shares motivations with related fields, such as the mining of interpretable characteristic rules or sub-group discovery \cite{Bach-subgroup,Bostrom}.

Similar efforts have been made to improve the human interpretability of prototypes for other data modalities. For example, in image analysis, \textit{prototypical parts networks} were introduced to identify characteristic patches rather than complete images \cite{chen2019protpnet}. Similarly, for text classification, adaptations of these networks can highlight the most important phrases instead of complete text paragraphs \cite{plucinski2021prototext}. The concept has also been applied to time-series, where prototypes are defined as representative shapes of important segments~\cite{bobek}. However, for tabular data, the decomposition of prototypes into meaningful parts remains underexplored.

To bridge this gap, we propose a method to identify the most important features within prototypes. Our approach applies a model-agnostic explanation method to compute feature importance for the black-box model, providing a more refined perspective than existing techniques. These identified feature subsets can be used for local and global explanations, helping users better interpret the provided results.

Our approach uses feature importance in two ways. First, we identify \textit{alike parts} by highlighting the most informative overlapping features between a predicted instance and its nearest prototype. This method directs the user's attention to a limited set of key features when interpreting a local model prediction. Second, we incorporate feature importance into the prototype selection objective function (used within the prototype generation algorithms) to promote diversity. This diversity aids in identifying meaningful alike parts. These strategies balance interpretability and diversity, enhancing both local explanations and the prototype selection process. The methods are evaluated on benchmark datasets.

This paper significantly extends our preliminary work \cite{KarolczakS25}. The new contributions include: (1) evaluating multiple prototype generation algorithms alongside a new evaluation function; (2) exploring a larger set of feature importance algorithms and feature selection operators for the final subsets; and (3) conducting a more extensive experimental analysis, assessing the impact of the proposed method's individual components and parameters.

\section{Related Work}

Below, we briefly review relevant work on prototype explanations and then on evaluating feature importance in black box models.

We consider a dataset $\mathcal{S}$ composed of $n$ learning examples, formally expressed as $\mathcal{S} = {(\mathbf{x}_i, y_i)}_{i=1}^{n}$. In this formulation, $\mathbf{x}_i \in \mathbb{X}^d$ is a feature vector that spans $d$ dimensions, and $y_i \in \mathcal{Y}$ is its associated label. The data format assumed throughout this work is tabular, presented as feature-value pairs. Central to our approach is a trained classifier $h$, which works as a black-box model for prediction. This classifier $h$ maps an input instance $\mathbf{x}_i$ to a predicted label $\hat{y}_i$, i.e. as the mapping: $h(\mathbf{x}_i)\mapsto\hat{y}_i$.

A \textit{prototype} as an instance chosen from the dataset that serves as a representative example, specifically $(\mathbf{x}_j, \hat{y}_j)$, where $\hat{y}_j$ is the class assignment yielded by the classifier $h$. The resulting collection of prototypes, denoted $\mathcal{P}$, constitutes a restricted subset of the entire dataset $\mathcal{S}$. Thus, $\mathcal{P} = \{(\mathbf{x}_j, y_j)\}_{j=1}^{m}$, where the condition $m \ll n$ ($\mathcal{P} \subset \mathcal{S}$) ensures that the final number of prototypes remains significantly smaller than the initial size of the training dataset.

Various methods for creating prototype explanations have been proposed up to now. Some methods rely on techniques such as vector quantization and kernel functions~\cite{schleif2011kernel}, sharing conceptual links with KNN-based approaches. For example, IKNN\_PSLFW~\cite{zhang2022distantprotos} first divides the data into class-specific groups before selecting prototypes that are maximally distant from instances belonging to other classes. A critical challenge for many  algorithms is their reliance on standard distance measures computed directly in the original feature space. This mandates a careful definition of similarity that can handle diverse data types (binary, numerical, categorical) and maintain robustness against scaling differences among features.

To overcome the limitations of distances in the raw feature space, recent approaches have focused on locating prototypes based on their proximity within a new representational space linked directly to the predictions of the black-box model. This shift is exemplified by the tree-space prototypes developed to explain ensemble models. They use a different view on evaluating classified instances based on their predictions; in the case of tree ensembles, the similarity of examples is determined by the degree of common decision leaves in the component trees. In this new space of this similarity, various methods of selecting representatives can be applied, drawing inspiration mainly from clustering algorithms. The initial algorithm in this line of work, SM-A \cite{tan2020tsp}, identifies prototypes as medoids in this novel prediction space, although it requires the user to pre-specify the desired number of prototypes. A later approach, A-PETE \cite{karolczak2024apete}, subsequently extends this process by providing a mechanism for autonomous prototype selection, using greedy medoid algorithms optimized with a specialized evaluation function (see later definitions in chapter \ref{subsec:augmented-objective}). The experiments  show satisfactory agreement between the surrogate classifier using 1-NN nearest prototype similarity and the prediction of the approximated tree ensemble \cite{karolczak2024apete}.

Despite the aforementioned research on prototypes for tabular data, surprisingly few studies have addressed the optimal manner of presenting these prototypes to end-users. Although visualizations like 2D scatter plots or self-organizing maps have been proposed in works like \cite{biehl2016prototypebasedmodels}, these are typically only suitable for low-dimensional data and ultimately fail to direct the user's focus toward the most critical parts of the prototype.

The concept of exposing only a subset of an instance or prototype to the user to reduce cognitive load is not novel. This practice is widely adopted in image processing, notably by ProtoPNet~\cite{chen2019protpnet} and its subsequent extensions, which present specific regions or parts of images. This idea has also been adapted for tabular data, as seen in applications such as medical laboratory testing~\cite{karolczak2025medic}. However, these existing approaches typically necessitate the use of highly specific deep learning architectures. Consequently, they are not readily applicable to other powerful predictive models, such as tree ensembles, which frequently represent the state-of-the-art solution for many real-world problems.

When assessing the importance of features and their selection, it should be noted that this issue has been extensively studied in KDD and ML for decades, but mainly at the level of assessing the global importance of features in the data itself and their interdependence with the target decision, see surveys \cite{Liu2018,Motoda-book}. Usually, statistical or other methods derived from information theory and wrappers around classifiers are exploited. 

However, in our approach, we need specialized methods for assessing the importance of features for ML model predictions for individual instances. According to reviews \cite{bodria2023benchmarking}, this can be achieved by a number of proposals based on different principles. Here we direct our interest to the popular SHAP family of methods for approximate Shapley values, such as Tree SHAP~\cite{lundberg2017shap}, which generates a local explanation in the form of a vector whose length corresponds to the number of features. The values within this vector attribute an importance score to each individual feature, which helps users understand the model's behavior for a specific instance. The spectrum of feature importance techniques is broad. Some methods exploit the inherent structure of the model itself. For example, Tree Interpreter~\cite{li2019treeinterpreter} capitalizes on the architecture of tree-based models, which allows it to run rapidly. In contrast, other techniques are entirely model-agnostic, relying solely on observing the impact of perturbations to the input data on the model's output decision. These methods, exemplified by LIME and Kernel SHAP~\cite{lundberg2017shap}, are computationally expensive and therefore operate very slowly, as they do not analyze the internal workings of the predictor at all.
 
Although numerous methods exist to assess the influence of features on predictions made by black-box models, these techniques have been underexploited in combination with prototype-based explanations. 

\section{Proposed Method}

This section outlines the main methodological contributions of this paper. Section~\ref{subsec:alike-parts} presents a method for guiding user attention in local explanations of decisions for an instance and its nearest prototype. Section~\ref{subsec:augmented-objective} then introduces a general approach to improving prototype-selection algorithms to produce prototypes that are better tailored to guide user attention. 

The introduced method is formalized in a quite general way and can be used with various algorithms for constructing the prototypes. For example, in our experiments, we will consider algorithms such as A-Pete or SM-A in relation to tree ensembles. 

\subsection{Identifying Alike Parts}
\label{subsec:alike-parts}

\begin{table*}[t!]
    \centering
    \caption{Demonstration of alike parts identification for a sample instance and its nearest prototype within the Apple Quality dataset. The table illustrates the process of generating the binary feature mask $\mathbf{m}$. The first and second rows display the raw feature-importance scores ($\phi$) for the instance and the prototype, respectively. The third row shows the resulting weight vector ($\mathbf{w}$), obtained using the Hadamard product as the similarity operator ($\circ$) on the normalized feature importance scores (as detailed in Section~\ref{subsec:alike-parts}). The final row, the binary mask ($\mathbf{m}$), depicts the most relevant shared features (here, where the weight exceeds the mean threshold), denoted by the value `1'.}
    \label{tab:mask-example}
    \begin{tabular}{lccccccc}
        \toprule
         & Size & Weight & Sweetness & Crunchiness & Juiciness & Ripeness & Acidity \\
        \midrule
        Instance & -2.77 & -1.08 & -1.72 & 1.38 & 0.19 & 3.65 & 0.31 \\
        Prototype & -0.97 & -0.20 & -3.07 & 0.00 & -0.52 & 3.16 & -0.52 \\
        Weights & 0.18 & 0.02 & 0.27 & 0.00 & 0.00 & 0.51 & 0.00 \\
        Mask & 1 & 0 & 1 & 0 & 0 & 1 & 0 \\
        \bottomrule
    \end{tabular}
\end{table*}

As noted by \cite{mastromichalakis2024prototypes}, a prototype can often be challenging to interpret as a whole (with all feature value pairs), which in turn complicates explaining the prediction of a black-box model. Individual features within the prototype may contribute unequally, some playing a important role in the prediction being interpreted, while others have little influence.

We introduce a method to pinpoint the most informative features that a predicted instance shares with its prototype, highlighting the focused feature subset for the user’s attention. These shared features, which we call \textit{alike parts}, consist of feature subsets whose importance remains consistently high in both the instance and its nearest prototype.

\begin{algorithm}[!t]
\caption{Alike parts identification.}
\label{alg:alike-parts-identification}
    \textbf{Input:} classifier $h$, instance to be explained $\mathbf{x}_i$, and its nearest prototype $\mathbf{p}_j$
    
    $\mathbf{f}_x,\,\mathbf{f}_p \gets \text{get\_fi}(h, \mathbf{x}_i),\,\text{get\_fi}(h, \mathbf{p}_j)$
    
    \textbf{if} normalize\_similarity \textbf{then} normalize($\mathbf{f}_x,\,\mathbf{f}_p$)
    
    \textbf{if} ignore\_direction \textbf{then} $\mathbf{f}_x,\,\mathbf{f}_p \gets |\mathbf{f}_x|,\,|\mathbf{f}_p|$
    
    $\mathbf{w} \gets \mathbf{f}_x \circ \mathbf{f}_p$
    
    mask $\gets$ weights\_to\_mask($\mathbf{w}$)

    \textbf{Output:} mask
\end{algorithm}

For this aim, we propose alike part identification (see Algorithm~\ref{alg:alike-parts-identification}), which outputs a binary mask highlighting the features the user should focus on when comparing an instance $\mathbf{x}_i$ with its nearest prototype $\mathbf{p}_j$.

Identifying alike parts starts by computing \textit{feature importance scores} $\phi(h, \mathbf{x}_i^l)$ and $\phi(h, \mathbf{p}_j^l)$ for each feature $l \in \{1, \dots, d\}$ using the classifier $h$. The method assumes that each feature is assigned exactly one score, positive or negative, without any normalization requirement. Hence, any feature importance estimation technique can be employed. For random forests, for example, one may use Tree SHAP~\cite{lundberg2017shap} or Tree Interpreter~\cite{li2019treeinterpreter}. Note that this is the most computationally expensive step of the proposed method. Therefore, while any local feature importance estimation technique can be utilized, prioritizing one with high computational efficiency is advised.

Optionally, the user can ignore the direction of feature importance scores (`ignore\_direction') by taking absolute values or normalize the scores (`normalize\_similarity'). The ignoring direction is defined as the simple absolute value of the raw score:

\begin{equation}
    \hat{\phi}(h, \mathbf{x}_i^l) = |\phi(h, \mathbf{x}_i^l))|\,.
\end{equation}

Optional scores normalization is defined as follows:

\begin{equation}
    \hat{\phi}(h, \mathbf{x}_i^l) = \frac{(\phi(h, \mathbf{x}_i^l))^2}{\sum_{k=1}^{d} (\phi(h, \mathbf{x}_i^k))^2}\,.
\end{equation}

The alignment of feature importance between the instance and its prototype is expressed as the similarity operator ($\circ$) applied to their respective scores:

\begin{equation}
    \label{eq:weight}
    w^l =  \hat{\phi}(h, \mathbf{x}_i^l) \circ \hat{\phi}(h, \mathbf{p}_j^l)\,.
\end{equation}

In this work, we consider several options for the similarity operator~$\circ$. These operators are applied element-wise to produce the weight vector $\mathbf{w}$ (where $w^l = \hat{\phi}(h, \mathbf{x}_i^l) \circ \hat{\phi}(h, \mathbf{p}_j^l)$):
\begin{itemize}
    \item Hadamard (element-wise): $w^l = \hat{\phi}(h, \mathbf{x}_i^l) \cdot \hat{\phi}(h, \mathbf{p}_j^l)$
    \item Element-wise $l_1$: $w^l = 1 - |\hat{\phi}(h, \mathbf{x}_i^l) - \hat{\phi}(h, \mathbf{p}_j^l)|$
    \item Element-wise $l_2$: $w^l = 1 - (\hat{\phi}(h, \mathbf{x}_i^l) - \hat{\phi}(h, \mathbf{p}_j^l))^2$
\end{itemize}
We select these measures because they are widely used in loosely related works, in particular~\cite{chen2019protpnet}.

The computed weights are critical, as they quantify each feature's importance in driving the model's prediction for both the nearest prototype and the classified instance. Based on the  values of these weights, a subset of features is selected, i.e., constructed  a binary feature mask, $\mathbf{m} \in \{0,1\}^d$. This selection process can be carried out in various ways. Below we propose the following strategies based on different principles: 

\begin{enumerate}
    \item Mean thresholding: The simplest approach retains any feature $l$ whose corresponding weight $w^l$ exceeds the average weight across all $d$ features: 
    \begin{equation}
        m^l = \mathds{1}\left( w^l > \frac{1}{d} \sum_{k=1}^{d} w^k \right)\,.    
    \end{equation}
    \item Adaptive Top-k Selection: To achieve explanations with controlled sparsity, we explore the selection of the $k$ features that possess the highest weights. We specifically test two data-driven scales for $k$: selecting the top $k=\lceil \sqrt{d} \rceil$ features and selecting the top $k = \lceil \log(d) \rceil$ features (they are inspired by Breiman's research on feature selection in  Random Forest \cite{breiman2001rf})
\end{enumerate}

In addition, the kneed algorithm~\cite{satopaa2011kneed} was also evaluated to identify the optimal feature-set size. However, it was dismissed because its hyperparameter sensitivity resulted in unstable outcomes, making it unreliable.

Table~\ref{tab:mask-example} serves to illustrate the initial selection procedure using the mean-based threshold.

\subsection{Augmented Objective for Prototype Selection}
\label{subsec:augmented-objective}

The algorithms established for prototype selection, including those presented in \cite{tan2020tsp,karolczak2024apete}, frame the problem of determining representative data points as a variation of the problem of $k$-medoids. This typically relies on a greedy approximation approach. The conventional $k$-medoids objective is to minimize a distance function $d$ between each training example $\mathbf{x}_i$ and its nearest prototype $\mathbf{p}_j$. This objective is standardly expressed as:

\begin{equation}
    f(\mathcal{P}) = \sum_{i=1}^{|\mathcal{S}|} \min_{\mathbf{p}_j \in \mathcal{P}} d \left( \mathbf{x}_i, \mathbf{p}_j \right),
\end{equation}

\noindent where $|\mathcal{S}|$ denotes the cardinality of the training set. The specific distance function $d$ utilized depends on the underlying architecture of the model. For example, in neural networks, this may involve a dot product of trainable embeddings \cite{li2018protonetwork}, or, for tree ensembles, a specialized tree distance metric \cite{tan2020tsp,karolczak2024apete}.

To diversify the influence of features in different prototypes, we propose extending the objective function by incorporating a term based on the feature importance score, $fi$. This component is derived from the similarity operator ($\circ$) defined in Section~\ref{subsec:alike-parts} and applied to the pre-processed feature importance scores ($\hat{\phi}$).

The feature importance score component, $fi$, is defined as the sum of the resulting weights $w^l$ (from Equation~\ref{eq:weight}, where $w^l = \hat{\phi}(h, \mathbf{x}_i^l) \circ \hat{\phi}(h, \mathbf{p}_j^l)$) for the $l$-th feature of instance $\mathbf{x}_i$ and its closest prototype $\mathbf{p}_j$:

\begin{equation}
    fi(\mathbf{x}_i, \mathbf{p}_j) = \sum_{l=1}^{d} w^l = \sum_{l=1}^{d} \left( \hat{\phi}(h, \mathbf{x}_i^l) \circ \hat{\phi}(h, \mathbf{p}_j^l) \right).
\end{equation}

The specific functional form of $fi$ thus depends on the chosen configuration for the $\hat{\phi}$ scores (i.e., whether the direction is ignored or normalization is applied) and the selected similarity operator $\circ$ (i.e., the Hadamard product, element-wise $1-l_1$, or element-wise $1-l_2$).

The $fi$ scores across all $\mathbf{x}_i$ can be pre-calculated and stored efficiently before the optimization procedure begins. The reformulated objective function integrates the classical distance minimization with an adjustable term, weighted by the parameter $\beta$, which accounts for the defined feature importance score. The first term ensures that the dataset $\mathcal{S}$ is compactly covered by the prototypes, as each instance is assigned to nearest representative. The second $\beta$-weighted term promotes a wider representation of feature importance among the selected prototypes. The new optimization objective is formally presented as:

\begin{equation}
    \label{eq:new-optim}
    f(\mathcal{P}) = \sum_{i=1}^{|\mathcal{S}|} \min_{\mathbf{p}_j \in \mathcal{P}} \left( d \left( \mathbf{x}_i, \mathbf{p}_j \right) + \beta \cdot fi \left( \mathbf{x}_i, \mathbf{p}_j \right) \right)\,.
\end{equation}

This modification facilitates a more sophisticated global prototype selection process, where the parameter $\beta$ acts as a trade-off mechanism between the metric distance and feature importance alignment. This updated formulation ultimately leads to specilized prototype selection for identifying alike parts.

A key advantage of the proposed method is its inherent robustness when handling missing values. This resilience relies on the native capabilities of the underlying components. Since we employ prototype selection algorithms \cite{tan2020tsp,karolczak2024apete} using Random Forest (RF) \cite{breiman2001rf}, along with feature estimation methods such as Tree SHAP and Tree Interpreter, all of which natively support missing values, the entire method does not require additional preprocessing steps for missing data.

The augmented prototype selection scales quadratically, $\mathcal{O}(n^2)$, with the number of training instances $n$, driven by the pairwise distance and feature importance computations. Conversely, the online inference phase is highly efficient; it scales linearly, $\mathcal{O}(m)$, with the number of selected prototypes $m$. Because $m \ll n$, the resulting surrogate model maintains strictly bounded, compact memory usage.

\section{Experiments}
\label{sec:experiments}

The experiments aim (1) to demonstrate the usefulness of the proposed method in terms of identifying important subsets of features for selected datasets; (2) to study the impact of its components and hyperparameters.

This section first details the setup~(\ref{subsec:experimental-setup}), followed by a case study on alike parts~(\ref{subsec:case-study}), and concludes with an analysis of the surrogate model's performance~(\ref{subsection:comparison}) and the number of activating features~(\ref{subsec:activating-features}).
 
\subsection{Experimental Setup}
\label{subsec:experimental-setup}

The proposed framework was implemented in Python using standard machine learning libraries, including scikit-learn\footnote{\href{https://scikit-learn.org}{https://scikit-learn.org}} for model training, SHAP\footnote{\href{https://shap.readthedocs.io}{https://shap.readthedocs.io}} and TreeInterpreter\footnote{\href{https://github.com/andosa/treeinterpreter}{https://github.com/andosa/treeinterpreter}} for feature importance estimation, and NumPy\footnote{\href{https://numpy.org}{https://numpy.org}} for custom functions and vectorized operations. The compact, modular codebase and reproducible experiment scripts are publicly available on GitHub~\footnote{\href{https://github.com/jkarolczak/alike-parts}{https://github.com/jkarolczak/alike-parts}}. Empirically, for the evaluated datasets, the prototype selection pipeline executed in the order of seconds on a personal computer (Apple M2, 16GB RAM). The resulting 1-NN surrogate models maintain minimal memory usage and provide highly efficient inference, demonstrating execution times comparable to the standard baselines.

\subsubsection{Algorithms}
As detailed in Section~\ref{subsec:augmented-objective}, our augmented optimization method is broadly applicable to various algorithms. For this study, we specifically implemented this modification in three different prototype selection algorithms that rely on optimizing a specialized tree distance metric (based on the degree of achieving the same decision tree leaves) to explain the Random Forest (RF) ensemble~\cite{breiman2001rf}, a type of multiple classifier system known to perform well across many applications~\cite{wozniak}. These three algorithms were chosen because they represent diverse strategies for prototype selection within the tree distance space and could be easily augmented with the proposed extension of the objective function:

\begin{itemize}
    \item \textbf{G-KM:} Selects an equal, predetermined number of prototypes within each class (greedy $k$-Medoid approximation computed within classes).
    \item \textbf{SM-A} \cite{tan2020tsp}: Selects the prototype that yields the largest marginal improvement in the objective function across all classes.
    \item \textbf{A-PETE} \cite{karolczak2024apete}: Automates prototype selection based on relative improvement thresholds (for details, see \cite{karolczak2024apete}).
\end{itemize}

\subsubsection{Datasets}
\label{subsec:benchmark-datasets}

We utilize six benchmark datasets for evaluation, varying in dimensionality from 7 to 31 features. These datasets are chosen to represent two distinct feature importance profiles, enabling us to test the robustness of our selection methods under diverse feature sparsity conditions. Four datasets possess a subset of globally important features: Australia Rain, Breast Cancer, Diabetes, and Passenger Satisfaction. The remaining two, Apple Quality and Wine Quality, exhibit a high importance spread across most or all features.\footnote{All datasets sourced from \href{https://www.kaggle.com/datasets/}{https://www.kaggle.com/datasets/}. Respective paths: \href{https://www.kaggle.com/datasets/jsphyg/weather-dataset-rattle-package}{jsphyg/weather-dataset-rattle-package}, \href{https://www.kaggle.com/datasets/rahmasleam/breast-cancer}{rahmasleam/breast-cancer}, \href{https://www.kaggle.com/datasets/mathchi/diabetes-data-set}{mathchi/diabetes-data-set}, \href{https://www.kaggle.com/datasets/teejmahal20/airline-passenger-satisfaction}{teejmahal20/airline-passenger-satisfaction}, \href{https://www.kaggle.com/datasets/nelgiriyewithana/apple-quality}{nelgiriyewithana/apple-quality}, and \href{https://www.kaggle.com/datasets/taweilo/wine-quality-dataset-balanced-classification}{taweilo/wine-quality-dataset-balanced-classification}.}

\subsubsection{Metric}
\label{subsec:metric}

To evaluate the quality of the selected set of prototypes $\mathcal{P}$, we measure the fidelity of a surrogate model built from these prototypes. This approach is consistent with prior work \cite{tan2020tsp,karolczak2024apete}.

Following the evaluation procedure in \cite{tan2020tsp}, our \textbf{surrogate model} is a 1-Nearest Neighbor (1-NN) classifier that uses the selected set of prototypes $\mathcal{P}$ as its training data. The 1-NN classifier assigns a test instance $\mathbf{x}_{test}$ the label of its nearest prototype $\mathbf{p}_j \in \mathcal{P}$, where nearest is determined by the same specialized tree distance metric $d$ used in the selection process.

\textbf{Fidelity} is then defined as the accuracy of this 1-NN surrogate model in matching the predictions of the original black-box model on a hold-out test set $\mathcal{S}_{test}$. Fidelity measures how well the small set of prototypes can represent the underlying black-box model's decision boundaries for classifying new, unseen data.

\begin{figure*}[!t]
    {\centering \includegraphics[width=0.9\textwidth]{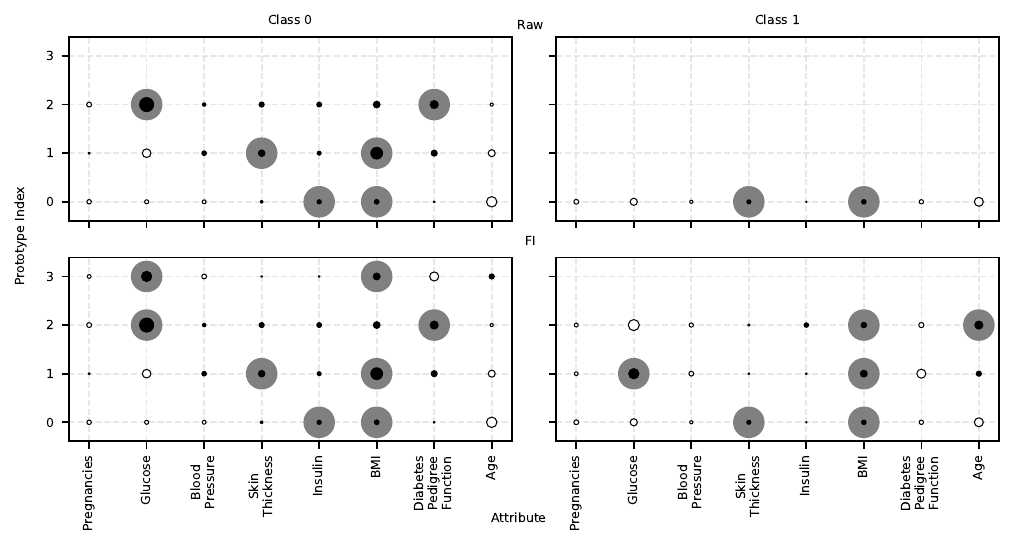}}
    
    {\small (a) Prototypes and features identified as important using the A-PETE algorithm and Tree Interpreter on the Diabetes dataset. The binary feature mask $\mathbf{m}$ was generated from the weight vector $\mathbf{w}$ by selecting $k=\lceil \log{d} \rceil$ features.}
    
    {\centering \includegraphics[width=\textwidth]{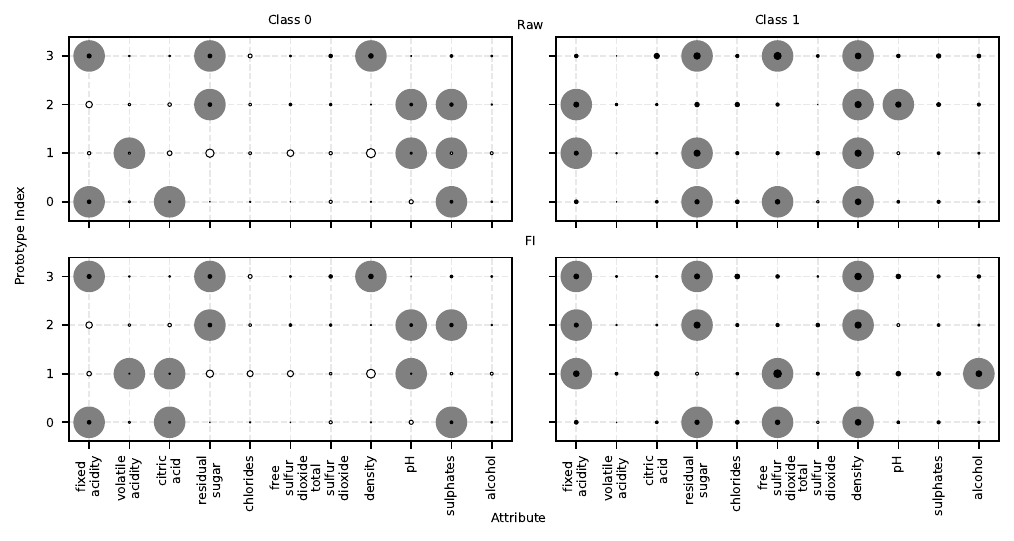}}
    
    {\small (b) Prototypes and features identified as important using the G-KM algorithm and Tree SHAP on the Wine Quality dataset. The binary feature mask $\mathbf{m}$ was generated from the weight vector $\mathbf{w}$ by selecting $k=\lceil \sqrt{d} \rceil$ features.}
    
    \caption{Comparison of prototypes (x-axis: prototype index) and important features (y-axis: feature index). The top row displays prototypes generated using the original raw algorithm, while the bottom row incorporates the augmented target function with feature importance (FI). The size of the inner circle represents the magnitude of the feature importance, and a gray highlight denotes features identified as important for a given prototype. The parameters employed in this analysis were selected to maximize fidelity, consistent with the results presented in Table~\ref{tab:fidelity}.}
    \label{fig:circles}
\end{figure*}

\subsection{Case Study: Analyzing Alike Parts}
\label{subsec:case-study}

\begin{table*}[t!]
    \caption{Example of alike parts identified for an instance and its nearest prototype using the A-PETE algorithm \protect\cite{karolczak2024apete}. The comparison illustrates feature selection results based on two distinct prototype optimization definitions: the original (raw) method and the Feature Importance (FI)-informed approach utilizing Tree SHAP. Features identified as alike between the explained instance and the prototype are distinguished by formatting: the FI-informed alike parts are displayed in \textbf{bold}, and alike parts selected in the prototypes from original (raw) problem formulation are \protect\underline{underlined}.}
    \label{tab:example-diabetes}
    \centering
    \begin{tabular}{llccccccccc}
    \hline
    class & type & Pregnancies & Glucose & BloodP. & SkinT. & Insulin & BMI & PedigreeF. & Age \\
    \hline
    \multirow{3}{*}{0} & instance & 6 & \underline{\textbf{102}} & 82 & 0 & 0 & 30.8 & \textbf{0.18} & \textbf{36} \\
    & prototype (FI) & 7 & \textbf{125} & 86 & 0 & 0 & 37.6 & \textbf{0.30} & \textbf{51} \\
    & prototype (Raw) & 7 & \underline{62} & 78 & 0 & 0 & 32.6 & 0.39 & 41 \\
    \hline
    \multirow{3}{*}{1} & instance & \underline{\textbf{8}} & \underline{\textbf{100}} & 74 & 40 & 215 & 39.4 & \textbf{0.66} & 43 \\
    & prototype (FI) & \textbf{9} & \textbf{152} & 78 & 34 & 171 & 34.2 & \textbf{0.89} & 33 \\
    & prototype (Raw) & \underline{9} & \underline{171} & 110 & 24 & 240 & 45.4 & 0.72 & 54 \\
    \hline
    \end{tabular}
\end{table*}

The mechanism for identifying alike parts is demonstrated as a pedagogical example in Table \ref{tab:mask-example}, which details the computation of weights from the feature importance vectors of an instance and its prototype. 

Furthermore, Table \ref{tab:example-diabetes} presents a comparison of the selected alike parts for two exemplary instances and its nearest prototype, contrasting the original (raw) and the Feature Importance (FI)-informed versions of the A-PETE algorithm on the Diabetes dataset. Integrating the feature importance component into A-PETE's optimization resulted in a noticeably different set of prototype selections than the raw algorithm when operating with a black-box RF~\cite{breiman2001rf}.

The prototype chosen by the raw A-PETE only identifies \textit{Glucose} as a feature important to both the instance and the prototype. In contrast, the FI-informed algorithm additionally highlights \textit{Diabetes Pedigree Function} and \textit{Age}. This outcome is consistent with established medical literature on diabetes risk factors \cite{kautzky2016diabetes}. This result illustrates our method's capacity to facilitate the discovery of more relevant and meaningful relationships between target instances and their representative prototypes.

Beyond this single-instance analysis, a visual comparison of the globally generated sets of prototypes and their selected important features for the Diabetes and Wine Quality datasets is presented in Fig.~\ref{fig:circles}. This figure directly contrasts the prototypes produced by the original (raw) algorithms with those generated using the Feature Importance (FI)-informed approach. The visualization confirms that the FI-informed algorithm generates prototypes with greater diversification, effectively highlighting feature subsets that vary between the selected representatives. 

For example, Fig.~\ref{fig:circles}a illustrates that the \textit{Age} feature was selected as an important feature in at least one prototype only when FI was incorporated into the objective function. Fig.~\ref{fig:circles}b depicts a comparable scenario for the Wine Quality dataset, where \textit{alcohol} is selected as important. This identical phenomenon was also observed for the Australia Rain and Breast Cancer datasets, where certain features were deemed important exclusively by the FI-informed version of the algorithm. Furthermore, augmenting the G-KM algorithm with FI -- an algorithm designed to find a predefined number of prototypes -- results in a more diversified set of prototypes, which is particularly observable in the prototypes selected for class~1 in Fig.~\ref{fig:circles}b.

\subsection{Surrogate Model Performance}
\label{subsection:comparison}

\begin{table*}[t]
    \centering
    \caption{Comparison of fidelity of prototypes in 1-NN search selected using three different configurations for the feature importance (FI) term in the augmented objective function. Interpretable models (NB: Naive Bayes, LR: Logistic Regression, DT: Decision Tree) trained on black-box predictions are included as baselines.}
    \label{tab:fidelity}
    \begin{tabular}{l ccc | l ccc}
        \toprule
        \multirow{2}{*}{Dataset} & \multicolumn{3}{c|}{Baseline Fidelity} & & \multicolumn{3}{c}{Prototype Fidelity} \\
        \cmidrule{2-4} \cmidrule{6-8}
        & NB & LR & DT & Method & Tree SHAP & Tree Interpreter & Raw \\
        \midrule
        \multirow{3}{*}{Apple Quality} 
        & \multirow{3}{*}{0.811} & \multirow{3}{*}{0.809} & \multirow{3}{*}{0.833} & A-Pete & 0.864 & 0.858 & 0.787 \\
        & & & & G-KM & 0.949 & 0.950 & 0.932 \\
        & & & & SM-A & 0.611 & 0.846 & 0.787 \\
        \midrule
        \multirow{3}{*}{Australia Rain} 
        & \multirow{3}{*}{0.859} & \multirow{3}{*}{0.932} & \multirow{3}{*}{0.835} & A-Pete & 0.892 & 0.916 & 0.930 \\
        & & & & G-KM & 0.935 & 0.963 & 0.956 \\
        & & & & SM-A & 0.887 & 0.899 & 0.893 \\
        \midrule
        \multirow{3}{*}{Breast Cancer} 
        & \multirow{3}{*}{0.947} & \multirow{3}{*}{0.939} & \multirow{3}{*}{0.921} & A-Pete & 0.965 & 1.000 & 1.000 \\
        & & & & G-KM & 1.000 & 1.000 & 1.000 \\
        & & & & SM-A & 0.851 & 0.956 & 1.000 \\
        \midrule
        \multirow{3}{*}{Diabetes} 
        & \multirow{3}{*}{0.890} & \multirow{3}{*}{0.890} & \multirow{3}{*}{0.818} & A-Pete & 0.864 & 0.896 & 0.903 \\
        & & & & G-KM & 0.955 & 0.916 & 0.909 \\
        & & & & SM-A & 0.948 & 0.831 & 0.883 \\
        \midrule
        \multirow{3}{*}{Passenger Satisfaction} 
        & \multirow{3}{*}{0.913} & \multirow{3}{*}{0.894} & \multirow{3}{*}{0.885} & A-Pete & 0.962 & 0.933 & 0.962 \\
        & & & & G-KM & 0.952 & 0.952 & 0.952 \\
        & & & & SM-A & 0.856 & 0.856 & 0.942 \\
        \midrule
        \multirow{3}{*}{Wine Quality} 
        & \multirow{3}{*}{0.705} & \multirow{3}{*}{0.655} & \multirow{3}{*}{0.705} & A-Pete & 0.786 & 0.769 & 0.760 \\
        & & & & G-KM & 0.867 & 0.829 & 0.798 \\
        & & & & SM-A & 0.786 & 0.724 & 0.769 \\
        \bottomrule
    \end{tabular}
\end{table*}

\begin{figure*}[!t]
    \centering
    \includegraphics[width=0.45\textwidth]
    {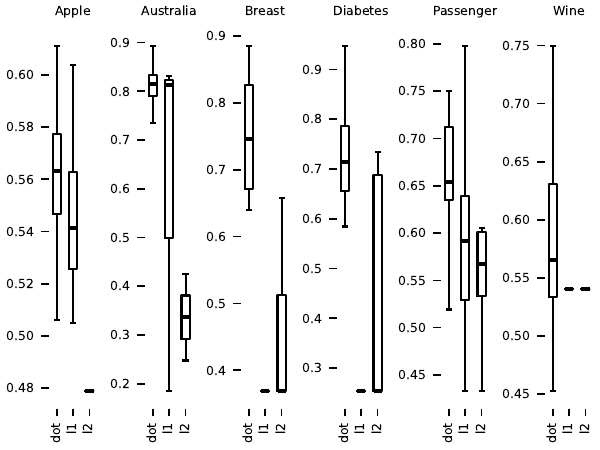}\hspace{0.5cm}
    \includegraphics[width=0.45\textwidth]
    {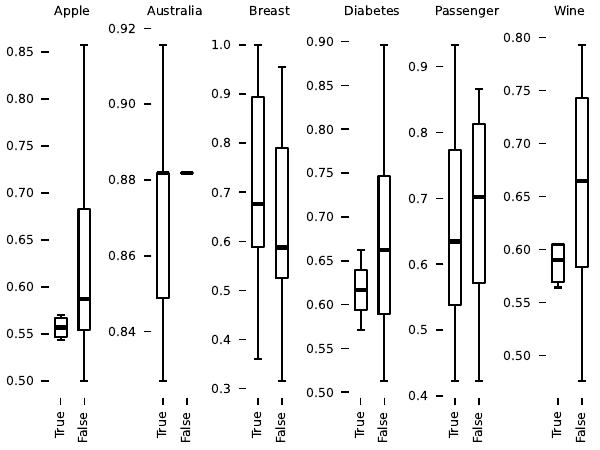}\\
    {\small (a)\hspace{8cm}(b)}
    \caption{Fidelity comparison of prototypes found with augmented objective across various configurations. The boxplots illustrate the distribution of fidelity across datasets. Subfigure (a) compares impact of similarity metrics for SM-A and Tree SHAP. Subfigure (b) compares the effect of ignoring the direction of feature importance (applying absolute value) for APete and Tree Interpreter.}
    \label{fig:boxplots}
\end{figure*}

The predictive performance of the surrogate 1-NN model, which is constructed from the generated prototypes, is quantified using fidelity. This evaluation is consistent with the established methodology found in related literature \cite{tan2020tsp,karolczak2024apete}. This surrogate model classifies instances from a hold-out test set by employing a 1-nearest neighbor (1-NN) search within the subset of selected prototypes. Our primary focus here is to quantify the impact of our modified prototype selection approach, which integrates feature importance, on the resulting model's accuracy. All hyperparameter settings were determined through extensive hyperparameter tuning using Bayesian optimization implemented via Optuna\footnote{\href{https://optuna.org}{https://optuna.org}} and the Tree-structured Parzen Estimator \cite{watanabe2025parzen}.

The maximal fidelity results for both the original (raw) and the feature importance (FI)-augmented prototype selection algorithms are summarized in Table~\ref{tab:fidelity}. As baselines, we evaluate three interpretable proxy explainers: Naive Bayes, Logistic Regression, and a Decision Tree (capped at depth 15 to balance capacity and interpretability). These are trained on the black-box Random Forest predictions to approximate its decision boundaries, using mean imputation for missing data.

It can be observed that integrating feature importance into the objective function yields highly competitive fidelity. A Wilcoxon signed-rank test comparing the original prototype selection against the proposed feature-informed objective indicates a positive trend toward improved fidelity ($p \approx 0.068$). Furthermore, evaluating the proposed prototype-based 1-NN explainers against the standard predictive baselines demonstrates a significant improvement in fidelity ($p < 0.05$).

\subsection{Sensitivity Analysis}

Fig.~\ref{fig:boxplots} examines the influence of two key hyperparameters on the fidelity of the resulting model. Fig.~\ref{fig:boxplots}a shows that for Tree SHAP and SM-A, the Hadamard product as a similarity metric works best across all datasets. In general, a specific hyperparameter value can show a global preference by consistently yielding higher fidelity across all evaluated datasets. However, as observed in Fig.~\ref{fig:boxplots}b for the `ignore\_direction' hyperparameter, while there is occasionally some trend, tailored tuning remains necessary, as there are usually datasets that do not match the global trend. The complete set of figures for all hyperparameters is available in the electronic appendix\footnote{\href{https://jkarolczak.github.io/appendix-alike-parts}{https://jkarolczak.github.io/appendix-alike-parts}}.

\subsection{Activating Features}
\label{subsec:activating-features}

\begin{figure}[t!]
    \centering
    \includegraphics[width=0.405\textwidth]{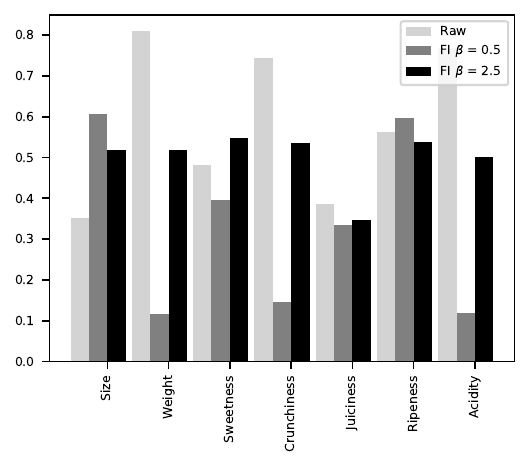}
    \caption{The comparison of the frequency of feature highlighting between the original (raw) and Feature Importance (FI)-informed strategies (with two beta values) for Apple Quality dataset. The prototypes were found using the A-PETE algorithm and Tree SHAP, with the mean threshold as the masking strategy.}
    \label{fig:activating-features}
\end{figure}

\begin{figure*}[!t]
    \centering
    \includegraphics[width=0.85\textwidth]
    {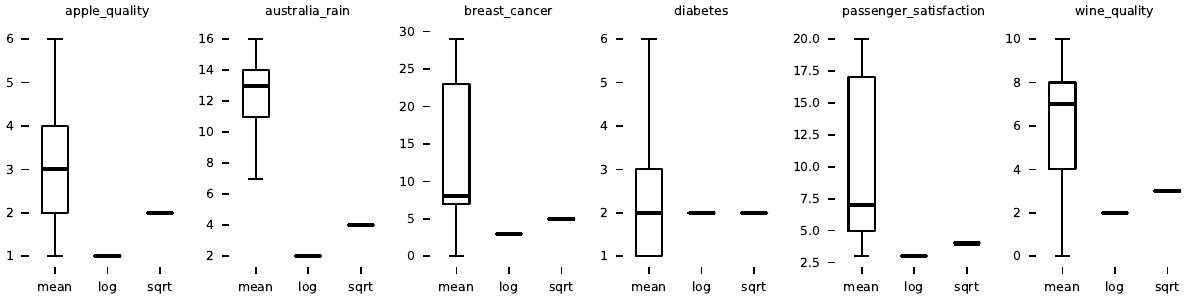}
    \caption{Statistics for the number of features identified as alike parts across all datasets. The results were compiled from all algorithm and masking combinations, using the specific hyperparameters that maximized test set fidelity (as reported in Table~\ref{tab:fidelity}).}
    \label{fig:number-of-features}
\end{figure*}

The proposed approach was validated by analyzing feature activation frequencies on the test sets. Figure \ref{fig:activating-features} shows that these frequencies differ notably between the original (raw) and Feature Importance (FI)-informed strategies. For the Apple Quality dataset (A-PETE with Tree SHAP), activation frequencies vary less when $\beta=2.5$ compared to $\beta=0.5$. This reduced variance demonstrates that the FI-informed objective successfully promotes feature diversity. Explicitly penalizing repeated feature usage forces the selection of a more diverse set of prototypes, ensuring broader coverage of feature influence.

Fig.~\ref{fig:number-of-features} provides a quantitative comparison of the number of features included in the identified alike parts across various experimental configurations, including different datasets, prototype selection algorithms, feature importance estimators, and feature masking strategies.

A clear trend emerges regarding the length of the resulting feature subset: the mean thresholding strategy typically generates the longest alike parts. This is followed sequentially by the two top-$k$ approaches: $k=\lceil\sqrt{d}\rceil$, and finally the most restrictive, $k=\lceil\log{d}\rceil$.

Crucially, the mean thresholding method offers the greatest inherent flexibility because its threshold is data-dependent, allowing it to dynamically adapt the resulting length of the alike part to the specific characteristics of each instance-prototype pair.

\section{Discussion}

This paper addresses a key limitation of prototype-based explanations: their failure to highlight which specific features drive the similarity  of a prototype for a given instance. We introduce the concept of alike parts by integrating feature importance directly into the prototype selection objective. This dual approach enhances local explanations by focusing user attention while simultaneously improving the global set of prototypes by promoting feature diversity.

A primary finding is that augmenting the prototype selection objective with a feature importance term, a move intended to boost feature diversity, does not necessarily degrade model fidelity. As shown in Section~\ref{subsection:comparison}, this augmentation frequently preserved or even increased the fidelity of the 1-NN surrogate model. This suggests that seeking a diverse set of feature attributions can be complementary to the distance-based coverage requirement. This global optimization for feature diversity also yielded qualitatively superior local explanations. For example, in the Diabetes case study (Section~\ref{subsec:case-study}), the feature-informed method successfully identified prototypes that share clinically relevant factors such as \textit{Age} and \textit{Pedigree Function}, connections that the baseline distance-only method missed.

A key aspect of our framework is the adaptability of the augmented objective function (Equation~\ref{eq:new-optim}). The hyperparameter $\beta$ (analyzed in Section~\ref{subsec:activating-features}) provides explicit control over the trade-off between traditional distance-based coverage ($d$) and feature importance alignment ($fi$). Although our experiments utilized a tree-specific metric for Random Forest, the framework's modular, model-agnostic design allows for the easy substitution of these components. For instance, it can be applied to neural networks by defining $d$ as the $l_2$ distance within the penultimate layer's embedding space and deriving $fi$ using architecture-specific estimators like Deep SHAP or Integrated Gradients. Furthermore, the proposed diversity measure readily adapts to conceptually different paradigms. In clustering algorithms, it can be integrated directly into the distance metric. For instance selection, retention criteria can be modified to preserve distinct feature-level explanations. In gradient-based learning, it can serve as a regularization term in the loss function. This design allows us to extend our framework to various architectures simply by substituting the distance function $d$ and the importance estimator $fi$.

Rather than modifying data, alike parts guide user attention toward shared rationale, while unshared features indicate where local explanations diverge. Eliminating unselected features alters the prototype's traversal through the Random Forest; missing values can redirect it into different leaves, potentially changing its class assignment. Retaining the full feature set ensures the explanation remains consistent with the model's internal logic.

Crucial future work is to empirically validate our central hypothesis: that alike parts improve user comprehension. This claim, currently supported by proxy metrics, must be tested directly through formal human-computer interaction studies using user-centric tasks. A significant technical extension would be to adapt the framework to other black-box models, such as deep neural networks, by substituting tree-specific components with counterparts such as latent space distances and gradient-based attribution methods for $fi$.

\begin{acknowledgment}
% This research was funded in part by National Science Centre, Poland OPUS grant no. 2023/51/B/ST6/00545 and in part by PUT SBAD 0311/SBAD/0770 grant.

The research of Jerzy Stefanowski was funded by National Science Centre, Poland OPUS grant no. 2023/51/B/ST6/00545. The research of Jacek Karolczak was funded by the PUT SBAD 0311/SBAD/0770 grant. For the purpose of Open Access, the author has applied a CC-BY public copyright licence to any Author Accepted Manuscript (AAM) version arising from this submission.

\end{acknowledgment}

\bibliography{amcs}

@inproceedings{karolczak2024apete,
    author = {Karolczak, Jacek and Stefanowski, Jerzy},
    title = {\text{A-PETE}: Adaptive Prototype Explanations of Tree Ensembles},
    booktitle = {Progress in Polish Artificial Intelligence Research},
    volume = {5},
    pages = {2-8},
    year = {2024},
}

@inproceedings{tan2020tsp,
    author = {Tan, Sarah and Soloviev, Matvey and Hooker, Giles and Wells, Martin T.},
    title = {Tree Space Prototypes: Another Look at Making Tree Ensembles Interpretable},
    year = {2020},
    isbn = {9781450381031},
    doi = {10.1145/3412815.3416893},
    booktitle = {Proc. of ACM FODS},
    pages = {23–34},
    numpages = {12}
}

@inproceedings{lundberg2017shap,
    author = {Lundberg, Scott M. and Lee, Su-In},
    title = {A unified approach to interpreting model predictions},
    year = {2017},
    booktitle = {Proc. of NIPS 2017},
    pages = {4768–4777},
    doi = {10.5555/3295222.3295230}
}

@inproceedings{li2018protonetwork,
    author = {Li, Oscar and Liu, Hao and Chen, Chaofan and Rudin, Cynthia},
    title = {Deep learning for case-based reasoning through prototypes: a neural network that explains its predictions},
    year = {2018},
    booktitle = {Proc. of AAAI 2018},
    articleno = {432},
    numpages = {8},
    doi = {10.5555/3504035.3504467}
}

@inproceedings{mastromichalakis2024prototypes,
    author = {Menis Mastromichalakis, Orfeas and Filandrianos, Giorgos and Liartis, Jason and Dervakos, Edmund and Stamou, Giorgos},
    title = {Semantic Prototypes: Enhancing Transparency without Black Boxes},
    year = {2024},
    doi = {10.1145/3627673.3679795},
    booktitle = {Proc. of ACM CIKM 2024},
    pages = {1680–1688},
}

@inproceedings{chen2019protpnet,
    author = {Chen, Chaofan and Li, Oscar and Tao, Chaofan and Barnett, Alina Jade and Su, Jonathan and Rudin, Cynthia},
    title = {This looks like that: deep learning for interpretable image recognition},
    year = {2019},
    booktitle = {Proc. of NIPS 2019},
    articleno = {801},
    numpages = {12}
}

@InProceedings{plucinski2021prototext,
    author = "Pluci{\'{n}}ski, Kamil and Lango, Mateusz and Stefanowski, Jerzy",
    title = "Prototypical Convolutional Neural Network for a Phrase-Based Explanation of Sentiment Classification",
    booktitle = "Proc. of ECML PKDD 2021",
    year = "2021",
    pages = "457-472"
}

@article{bodria2023benchmarking,
    author = {Bodria, Francesco and Giannotti, Fosca and Guidotti, Riccardo and Naretto, Francesca and Pedreschi, Dino and Rinzivillo, Salvatore},
    year = {2023},
    title = {Benchmarking and Survey of Explanation Methods for Black Box Models},
    journal = {Data Mining and Knowledge Discovery},
    volume = {37},
    pages = {1719-1778},
    doi = {10.1007/s10618-023-00933-9}
}

@article{zhang2022distantprotos,
    title = {K-nearest neighbors rule combining prototype selection and local feature weighting for classification},
    journal = {Knowledge-Based Systems},
    volume = {243},
    pages = {108451},
    year = {2022},
    author = {Xin Zhang and Hongshan Xiao and Ruize Gao and Hongwu Zhang and Yu Wang},
}

@article{schleif2011kernel,
    author = {Schleif, F.-M. and Villmann, Thomas and Hammer, Barbara and Schneider, Petra},
    title = {Efficient Kernelized Prototype Based Classification},
    journal = {International Journal of Neural Systems},
    volume = {21},
    number = {06},
    pages = {443-457},
    year = {2011},
    doi = {10.1142/S012906571100295X},
}

@article{biehl2016prototypebasedmodels,
    author = {Biehl, Michael and Hammer, Barbara and Villmann, Thomas},
    title = {Prototype-based models in machine learning},
    journal = {WIREs Cognitive Science},
    volume = {7},
    number = {2},
    pages = {92-111},
    doi = {10.1002/wcs.1378},
    year = {2016}
}

@article{breiman2001rf,
    author = {Breiman, Leo},
    title = {Random Forests},
    year = {2001},
    issue_date = {October 1 2001},
    volume = {45},
    number = {1},
    journal = {Machine Learning},
    pages = {5–32},
}

@inproceedings{ribeiro2016lime,
    author = {Ribeiro, Marco Tulio and Singh, Sameer and Guestrin, Carlos},
    title = {"Why Should I Trust You?": Explaining the Predictions of Any Classifier},
    year = {2016},
    doi = {10.1145/2939672.2939778},
    booktitle = {Proc. of ACM SIGKDD 2016},
    pages = {1135–1144},
    numpages = {10}
}

@article{kautzky2016diabetes,
  title={Sex and Gender Differences in Risk, Pathophysiology and Complications of Type 2 Diabetes Mellitus},
  author={Alexandra Kautzky-Willer and J{\"u}rgen Harreiter and Giovanni Pacini},
  journal={Endocrine Reviews},
  year={2016},
  volume={37},
  pages={278-316},
}

@inproceedings{li2019treeinterpreter,
    author = {Xiao Li and Yu Wang and Sumanta Basu and Karl Kumbier and Bin Yu},
    title = {A Debiased MDI Feature Importance Measure for Random Forests},
    booktitle = {Advances in Neural Information Processing Systems},
    year = {2019},
    volume = {32},
}

@misc{watanabe2025parzen,
      title={Tree-Structured Parzen Estimator: Understanding Its Algorithm Components and Their Roles for Better Empirical Performance}, 
      author={Shuhei Watanabe},
      year={2025},
      eprint={2304.11127},
      archivePrefix={arXiv},
      primaryClass={cs.LG},
}

@inproceedings{karolczak2025medic,
	author="{Karolczak, Jacek and Stefanowski, Jerzy}",
	title="{An Interpretable Prototype Parts-based Neural Network for Medical Tabular Data}",
	year="2025",
	booktitle="{Proc. of EXPLIMED at ECML PKDD 2025}",
    series="{CEUR WS}",
    volume="4059"
}

@inproceedings{satopaa2011kneed,
    author={Satopaa, Ville and Albrecht, Jeannie and Irwin, David and Raghavan, Barath},
    booktitle="{Proc. of ICDCS Workshops 2011}", 
    title={Finding a "Kneedle" in a Haystack: Detecting Knee Points in System Behavior}, 
    year={2011},
    volume={},
    number={},
    pages={166-171}
}

@article{TrustAI,
    author = {Kaur, Davinder and Uslu, Suleyman and Rittichier, Kaley J. and Durresi, Arjan},
    title = {Trustworthy Artificial Intelligence: A Review},
    year = {2022},
    issue_date = {February 2023},
    journal = {ACM Comput. Surv.},
    volume = {55},
    number = {2},
    idoi = {10.1145/3491209}
}

@ARTICLE{TrustXAI,
  author={Chamola, Vinay and Hassija, Vikas and Sulthana, A Razia and Ghosh, Debshishu and Dhingra, Divyansh and Sikdar, Biplab},
  journal={IEEE Access}, 
  title={A Review of Trustworthy and Explainable Artificial Intelligence (XAI)}, 
  year={2023},
  volume={11},
   pages={78994-79015},
   doi={10.1109/ACCESS.2023.3294569}
}

@misc{Bostrom,
    author = {Alkhatib, Amr and Boström, Henrik and Vazirgiannis, Michalis},
    year = {2024},
    month = {05},
    pages = {},
    title = {Explaining Predictions by Characteristic Rules},
}

@article{Bach-subgroup,
    author = {Bach, Jakob},
    title = {Subgroup Discovery with Small and Alternative Feature Sets},
    year = {2025},
    volume = {3},
    number = {3},
    doi = {10.1145/3725358},
    journal = {Proc. ACM Manag. Data 2025},
}

@inproceedings{KarolczakS25,
    author       = {Jacek Karolczak and Jerzy Stefanowski},
    title        = {This Part Looks Alike This: Identifying Important Parts of Explained Instances and Prototypes},
    booktitle    = {Proc. of the xAI 2025 Late-breaking Work, Demos and Doctoral Consortium at xAI},
    series       = "{CEUR WS}",
    volume       = {4017},
    pages        = {33--40},
    year         = {2025}
}

@book{Motoda-book,
    author = {Liu, Huan and Motoda, Hiroshi},
    year = {2000},
    month = {01},
    pages = {},
    title = {Feature Selection for Knowledge Discovery and Data Mining},
    publisher = {Kluwer Academic, USA},
    doi = {10.1007/978-1-4615-5689-3}
}

@article{Liu2018,
    author = {Li, Jundong and Cheng, Kewei and Wang, Suhang and Morstatter, Fred and Trevino, Robert P. and Tang, Jiliang and Liu, Huan},
    title = {Feature Selection: A Data Perspective},
    year = {2017},
    issue_date = {November 2018},
    volume = {50},
    number = {6},
    doi = {10.1145/3136625},
    journal = {ACM Comput. Surv.}
}

@article{wozniak,
    title = {A survey of multiple classifier systems as hybrid systems},
    journal = {Information Fusion},
    volume = {16},
    pages = {3-17},
    year = {2014},
    issn = {1566-2535},
    doi = {https://doi.org/10.1016/j.inffus.2013.04.006},
    author = {Michał Woźniak and Manuel Graña and Emilio Corchado}
}

@article{bobek,
    title = {TSProto: Fusing deep feature extraction with interpretable glass-box surrogate model for explainable time-series classification},
    journal = {Information Fusion},
    volume = {124},
    pages = {103357},
    year = {2025},
    issn = {1566-2535},
    doi = {https://doi.org/10.1016/j.inffus.2025.103357},
    author = {Szymon Bobek and Grzegorz J. Nalepa},
}

@article{stepka2024multi,
    author = {Stepka, Ignacy and Lango, Mateusz and Stefanowski, Jerzy},
    title = {A Multi–Criteria Approach for Selecting an Explanation from the Set of Counterfactuals Produced by an Ensemble of Explainers},
    year = {2024},
    volume = {34},
    number = {1},
    doi = {10.61822/amcs-2024-0009},
    journal = {International Journal of Applied Mathematics and Computer Science},     
    pages = {119–133},
}

\begin{biography}[author-jk]{Jacek Karolczak} is a PhD student in the Institute of Computing Science, Poznan University of Technology, where he also received his MSc and BSc degrees in Artificial Intelligence. His research interests focus on interpretable machine learning, including both ante-hoc and post-hoc approaches, with a particular emphasis on prototype-based explanations and machine learning for data streams with concept drift.
\end{biography}

\begin{biography}[author-js]{Jerzy Stefanowski} is a full professor in the Institute of Computing Science of the Poznan University of Technology and a member of the Polish Academy of Sciences. His research interests include machine learning, data mining and intelligent decision support, in particular rule induction, ensembles, class imbalance, concept drift, classification of data streams and explainable AI.
\end{biography}

% \makeinfo

\end{document}